\documentclass{article}

\usepackage{booktabs}
\usepackage{multirow}
\usepackage{xcolor}

\usepackage{CJKutf8}
\usepackage{longcat_style}
\usepackage{adjustbox}
\usepackage[utf8]{inputenc} %
\usepackage[T1]{fontenc}    %
\usepackage{newunicodechar}
\usepackage{hyperref}       %
\usepackage{xcolor}
\usepackage[normalem]{ulem} %
\hypersetup{
    colorlinks=true,      %
    linkcolor=blue,      %
    urlcolor=blue,       %
    citecolor=blue,      %
    linkbordercolor=blue, %
    urlbordercolor=blue,
    citebordercolor=blue,
    pdfborderstyle={/S/U/W 1}, %
}
\usepackage{float}
\usepackage{url}            %
\usepackage{booktabs}       %
\usepackage{amsfonts}       %
\usepackage{nicefrac}       %
\usepackage{microtype}      %
\usepackage{lipsum}		%
\usepackage{graphicx}
\usepackage{natbib}
\usepackage{doi}
\usepackage{amsmath}
\usepackage{amssymb} %
\usepackage{xspace}
\usepackage{enumitem}
\usepackage{multirow}
\usepackage{subcaption} 
\usepackage{makecell}
\usepackage{hyperref, cleveref}
\usepackage{pifont}
\usepackage[inkscapelatex=false]{svg}
\usepackage{caption}
\DeclareCaptionLabelSeparator{pipe}{ | }
\usepackage{tcolorbox}
\newtcolorbox{coloredquote}[1][]{
    colback=green!5!white,  
    colframe=green!70!black, 
    boxrule=2pt,
    arc=7pt,
    left=6pt,
    right=6pt,
    top=4pt,
    bottom=4pt,
    title=#1
}

\usepackage{xcolor}

\usepackage{subcaption}
\usepackage{graphicx}
\usepackage{amsmath}
\usepackage{wrapfig}
\usepackage{subcaption}     

\setlist[itemize]{leftmargin=*}
\setlist[enumerate]{leftmargin=*}
\setlist[description]{leftmargin=*}

\title{Multi-Objective and Mixed-Reward Reinforcement Learning via Reward-Decorrelated Policy Optimization}

\author{
Yang Bai\footnotemark[1], Kaiyuan Liu\thanks{Equal Contribution}, Ziyuan Zhuang, Jiahong Zhou, Rongxiang Weng\thanks{Corresponding author: wengrongxiang@meituan.com} \\ \textbf{Xin Chen, Jingang Wang, Xunliang Cai} \\
    Meituan, China \\
	\texttt{\{baiyang28, liukaiyuan07\}@meituan.com} \\
}

\renewcommand{\headeright}{\raisebox{-0.2\height}{\includegraphics[height=1.8em]{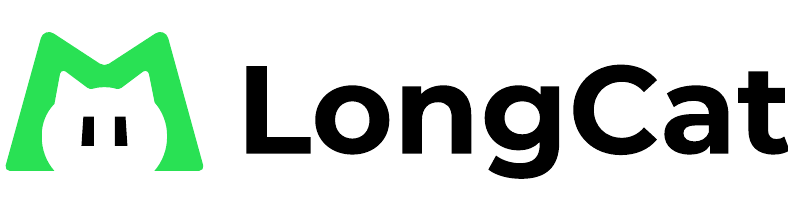}}}
\renewcommand{\shorttitle}{Reward-Decorrelated Policy Optimization}

\begin{document}
\maketitle
\setcounter{footnote}{0}

\begin{abstract}

Complex reinforcement learning environments frequently employ multi-task and mixed-reward formulations. In these settings, heterogeneous reward distributions and correlated reward dimensions often destabilize the construction of scalar advantages. To address these challenges, we propose Reward-Decorrelated Policy Optimization (RDPO), a reward-processing method designed to explicitly target both failure modes. RDPO first utilizes Magnitude-Aware Quantile normalization to stabilize prompt-level advantage allocation across binary, fractional, and continuous rewards. It then applies Mahalanobis whitening within each active reward subspace to mitigate correlation redundancy prior to aggregation. When applied during the post-training of LongCat-Flash, RDPO enhances instruction following, writing quality, and robustness to hard prompts while remaining broadly competitive on reasoning and coding evaluations.

\end{abstract}

\section{Introduction}

This technical report presents the RDPO post-training experiments for LongCat-Flash. We consider a standard yet challenging reinforcement learning setting: a single training run incorporates multiple task types. Each task provides a distinct subset of reward signals, such as correctness, instruction following, rubric satisfaction, preference-model scores, and response length. Aggregating these heterogeneous signals into a single scalar advantage often causes training instability. This instability arises because the rewards exhibit varying scales, diverse distribution shapes, and non-trivial correlations.

RDPO mitigates this challenge through a lightweight, two-step reward processing pipeline. First, Magnitude-Aware Quantile Normalization makes prompt-level advantages more robust against binary rewards, ties, skewed distributions, and outliers. Second, Mahalanobis whitening reduces redundant variance among reward dimensions that co-occur within a given task. The remainder of this report outlines the methodology, training setup, reward design, and evaluation results.

\section{Method}
\label{Method}

\begin{table}[h]
\centering
\small
\setlength{\tabcolsep}{12pt}
\renewcommand{\arraystretch}{1.4}
\begin{tabular}{@{}lcc@{}}
\toprule
& \textbf{Reward Normalization} & \textbf{Reward Aggregation} \\
\midrule
\textbf{GRPO}~\citep{DBLP:journals/corr/abs-2402-03300} 
& \multicolumn{2}{c}{$\displaystyle r_{\text{sum}}^{(i,j)} = \sum_{k=1}^{n} r_k^{(i,j)}$} \\
& \multicolumn{2}{@{}c@{}}{\textcolor{gray}{\small Direct summation of raw rewards}} \\
\midrule
\textbf{GDPO}~\citep{DBLP:journals/corr/abs-2601-05242} 
& $\displaystyle A_k^{(i,j)} = \frac{r_k^{(i,j)} - \mu_k^{(i)}}{\sigma_k^{(i)}}$ 
& $\displaystyle A_{\text{sum}}^{(i,j)} = \sum_{k=1}^{n} A_k^{(i,j)}$ \\
& \textcolor{gray}{\small Z-score Normalization} & \textcolor{gray}{\small Summation} \\
\midrule
\textbf{RDPO (Ours)} 
& $\displaystyle A_k^{(i,j)} = \Phi^{-1}\!\left(u_k^{(i,j)}\right)$ 
& $\displaystyle A_{\text{sum}}^{(i,j)} = \sum_{k=1}^{n} W_k^{(i,j)}, \mathbf{W}^{(i,j)} = \boldsymbol{\Sigma}^{-1/2} \mathbf{A}^{(i,j)}$ \\
& \textcolor{gray}{\small Magnitude-Aware Quantile Normalization} & \textcolor{gray}{\small Mahalanobis whitening} \\
\bottomrule
\end{tabular}
\caption{\textbf{Comparison of reward processing methods.} 
GRPO sums raw rewards without normalization, obscuring relative performance variations across reward dimensions. 
GDPO applies independent Z-score normalization but remains vulnerable to prompt-level advantage domination and cross-dimensional correlations. 
Our RDPO combines \textit{Magnitude-Aware Quantile Normalization} for stable advantage allocation with \textit{Mahalanobis whitening} for correlation reduction within active reward subspaces.
Detailed analysis is provided in Section~\ref{idpo_method}.}
\label{table1}
\end{table}

\subsection{Background}

In real-world deployments, large language models (LLMs) must simultaneously optimize for multiple objectives, such as computational efficiency~\citep{DBLP:journals/corr/abs-2501-12599,DBLP:journals/corr/abs-2503-04697}, alignment with human preferences~\citep{DBLP:journals/corr/abs-1706-03741}, and prompt-specific constraints~\citep{DBLP:journals/corr/abs-2510-07743}. This inherent complexity has motivated recent advances in reinforcement learning (RL) for multi-task and mixed-reward settings~\citep{DBLP:journals/corr/abs-2601-05242,DBLP:journals/corr/abs-2506-16141,DBLP:journals/corr/abs-2505-15612}, wherein a single rollout can yield diverse, heterogeneous reward signals. Below, we briefly outline two approaches:

\paragraph{GRPO}

For a given prompt $i$ with $G$ rollouts, let the $j$-th rollout receive $n$ rewards, denoted as $r^{(i,j)}=(r_1^{(i,j)},\dots,r_n^{(i,j)})^T$. GRPO~\citep{DBLP:journals/corr/abs-2402-03300} aggregates mixed-reward feedback by summing the raw rewards prior to group-level normalization: $r_{\text{sum}}^{(i,j)}=\sum_{k=1}^{n}r_k^{(i,j)}$. While this straightforward strategy is effective when rewards are on comparable scales, it can obscure the contributions of individual reward dimensions when their scales and underlying distributions differ.

\paragraph{GDPO}

GDPO~\citep{DBLP:journals/corr/abs-2601-05242} addresses this heterogeneity by normalizing each reward dimension independently prior to aggregation, as summarized in Table~\ref{table1}.
Specifically, for the $k$-th reward dimension, it computes the advantage with dimension-level Z-score normalization.
The final scalar advantage is then obtained by summing these normalized dimensions and applying batch-wise normalization. Although this decoupled approach improves upon raw summation, it still treats each reward independently, leaving the method sensitive to non-Gaussian reward distributions and inter-reward correlations.

\subsection{Effective Information Efficiency}

We introduce \textit{Effective Information Efficiency} ($\eta_{\text{eff}}$) as a diagnostic messure to evaluate mixed-reward aggregation.
The metric captures two complementary aspects of a scalar mixed advantage: whether the aggregation balances weights across reward dimensions, and whether the aggregated reward contains redundant variation caused by correlated reward dimensions.
Formally, we decompose it as follows:
$$\eta_{\text{eff}} = \eta_{\text{proj}} \times \eta_{\text{corr}}.$$
This decomposition follows two basic desiderata for a useful mixed-reward advantage. First, each active reward dimension should contribute on a comparable standardized scale. Second, the summed signal should not repeatedly count the same underlying variation. Therefore, $\eta_{\text{eff}}$ serves as a method-agnostic diagnostic of aggregation quality.

The first term, $\eta_{\text{proj}}$, measures how closely the aggregation direction aligns with an equally weighted projection in the standardized reward space. Let $z_k = (r_k - \mu_k) / \sigma_k$ be the standardized reward, and let $\mathbf{1}$ denote the all-ones vector. For an arbitrary aggregation weight vector $\mathbf{w}$, we define:$$\eta_{\text{proj}}(\mathbf{w}) = \cos^{2}(\mathbf{w}, \mathbf{1}) = \frac{(\mathbf{w}^{T}\mathbf{1})^{2}}{n \cdot \|\mathbf{w}\|^{2}}.$$

The second term, $\eta_{\text{corr}}$, quantifies the amount of independent information retained after summing correlated, standardized rewards. Both positive and negative correlations imply dependence across reward dimensions. Thus, we compute this term using the element-wise absolute correlation matrix $|\Sigma_z|$:$$\eta_{\text{corr}} = \frac{n}{\mathbf{1}^{T}|\Sigma_{z}|\mathbf{1}}.$$

For the two-reward case with a Pearson correlation of $\rho$, this simplifies to:$$\eta_{\text{corr}} = \frac{2}{2+2|\rho|} = \frac{1}{1+|\rho|}.$$
Thus, any strong linear dependency, whether positive or negative, reduces the amount of effective independent information present in the summed advantage.

We now apply this metric to analyze various reward-processing strategies. In the case of GRPO, we can express each raw reward as $r_k = \mu_k + \sigma_k z_k$. Direct reward summation yields:$$\sum_{k=1}^{n} r_k = \sum_{k=1}^{n}\mu_k + \sum_{k=1}^{n}\sigma_k z_k.$$

The constant term $\sum_k \mu_k$ is removed by group-level advantage normalization. As a result, the effective aggregation direction in the standardized reward space is determined purely by the coefficients of $z_k$. Therefore, GRPO implicitly relies on the weight vector $\mathbf{w}_{\text{GRPO}} = (\sigma_1, \sigma_2, \dots, \sigma_n)^T$. This assigns disproportionately larger effective weights to reward dimensions with higher raw variances. Substituting this weight vector into $\eta_{\text{proj}}$ yields:$$\eta_{\text{proj}}(\mathbf{w}_{\text{GRPO}}) = \frac{(\sum_{k=1}^{n}\sigma_{k})^{2}}{n\sum_{k=1}^{n}\sigma_{k}^{2}}.$$

\begin{wrapfigure}[17]{t}{0.50\columnwidth}
    \centering
    \includegraphics[width=0.45\textwidth]{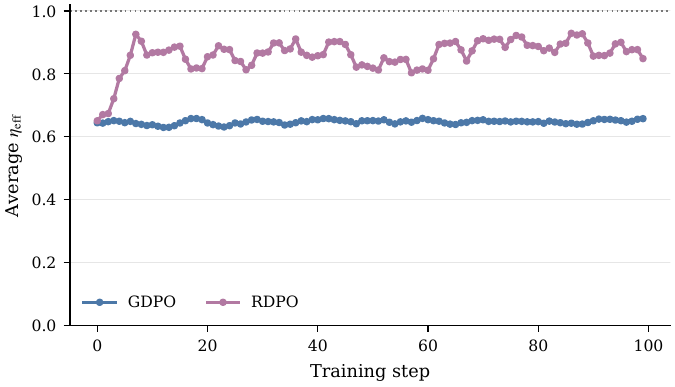}
    \caption{\textbf{Effective Information Efficiency across training.} For the four-task mixture, average $\eta_{\text{eff}}$ is computed by first evaluating each active reward subspace and then aggregating subspace values.}
    \label{eta_on_real_data}
\end{wrapfigure}

This formulation highlights how imbalances in reward scaling can diminish the effective contribution of certain dimensions. In contrast, GDPO first normalizes every reward dimension to $A_k = (r_k-\mu_k)/\sigma_k = z_k$ before summing them. Its aggregation direction is therefore $\mathbf{w}_{\text{GDPO}} = \mathbf{1}$, which perfectly aligns with the equally weighted reference direction and eliminates the variance-scaling loss captured by $\eta_{\text{proj}}$ at the reward-dimension level. In essence, GDPO does more than rescale rewards. It restores geometric consistency between the realized optimization direction and the intended preference direction, which underlies its effectiveness across mixed-reward landscapes.

However, Z-score normalization can still be unstable at the prompt level. When a prompt-level rollout group contains skewed rewards, binary outcomes, ties, or outliers, the normalized advantage mass can concentrate on a single rollout while the remaining rollouts receive near-zero or suppressed advantages. In such cases, the policy update is effectively driven by a few samples, making the equal-contribution assumption less reliable even after per-reward standardization. GDPO also assumes that reward dimensions can be aggregated independently, thereby failing to address the correlation loss captured by $\eta_{\text{corr}}$.

RDPO is designed to address the two failure modes measured by $\eta_{\text{eff}}$: Magnitude-Aware Quantile (MAQ) makes prompt-level normalized advantages less sensitive to heterogeneous reward scales and outliers, while Mahalanobis whitening reduces redundant variation among co-occurring reward dimensions within each active reward subspace. Further details regarding this mechanism are provided in the subsequent section. Figure \ref{eta_on_real_data} reports the average $\eta_{\text{eff}}$ across active task subspaces. As shown, RDPO maintains a higher effective information efficiency than the GDPO normalization baseline throughout training. Under the absolute-correlation definition above, an efficiency value of $1.0$ serves as an independent-reward reference baseline, and stronger dependencies monotonically reduce $\eta_{\text{corr}}$.

\subsection{Reward-Decorrelated Policy Optimization}
\label{idpo_method}

We first selected four representative tasks for our experiments: instruction following, general writing, mathematical reasoning, and code generation, all conducted within a unified post-training run. Each task incorporates two to three rewards; further details are provided in Section \ref{sec:reward_design}. This configuration exposes RDPO to subspaces containing two and three rewards, as well as a mixture of binary, discrete, and continuous reward distributions.

\subsubsection{Magnitude-Aware Quantile Normalization}

\begin{figure}[ht]
    \centering
    \includegraphics[width=0.9\textwidth]{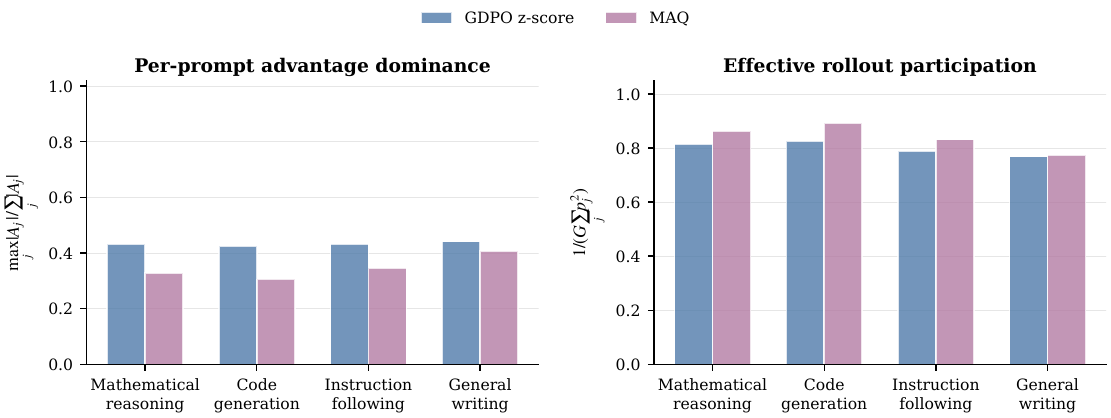}
    \caption{\textbf{MAQ stabilizes prompt-level advantage allocation.} We compare GDPO's Z-score normalization with MAQ across the four active task subspaces. The left panel measures per-prompt advantage domination, defined by the largest rollout's share of the absolute advantage mass. The right panel measures effective rollout participation, $1/(G\sum_j p_j^2)$, where $p_j=|A_j|/\sum_{\ell=1}^{G}|A_\ell|$ denotes the normalized absolute advantage mass of rollout $j$ within a prompt. MAQ consistently lowers domination and increases participation, indicating that it makes heterogeneous reward signals more comparable without allowing a single rollout to disproportionately dominate the prompt-level update.}
    \label{maq_compare}
\end{figure}

\textbf{The Problem:} The projection term $\eta_{\text{proj}}$ assumes that active reward dimensions contribute on a comparable, standardized scale prior to aggregation. Although GDPO attempts to satisfy this requirement by applying per-reward Z-score normalization, this linear transformation remains highly sensitive to the distribution shape of each prompt-level rollout group. To evaluate the stability of advantage allocation within each prompt, we compute prompt-level statistics and report the mean across each task subspace. Specifically, for normalized rollout advantages $\{A_j\}_{j=1}^{G}$, we use $p_j=|A_j|/\sum_{\ell=1}^{G}|A_\ell|$ to measure each rollout's share of the prompt-level absolute advantage mass. This gives two complementary diagnostics: advantage domination, $\max_j p_j$, measures whether a single rollout receives most of the update signal, while effective rollout participation, $1/(G\sum_j p_j^2)$, measures how evenly the advantage mass is distributed across the $G$ rollouts. This approach highlights typical prompt behavior rather than relying on a pooled distribution that could be heavily skewed by a few extreme groups. Because the underlying rewards in our setting can be binary, fractional, or continuous, phenomena such as skewed distributions, ties, and outliers can concentrate the majority of the normalized advantage mass onto a single rollout, even after Z-score normalization. Figure~\ref{maq_compare} illustrates this failure mode: GDPO frequently exhibits high per-prompt advantage concentration and lower effective rollout participation, indicating that the policy update may be driven by a small subset of rollouts rather than a stable, group-level comparison.

\textbf{The Solution:} To better satisfy the equal-contribution assumption underlying $\eta_{\text{proj}}$ under non-Gaussian reward groups, we propose \textbf{Magnitude-Aware Quantile (MAQ)} normalization. Z-score normalization provides the cleanest linear route to an equal-scale projection when the prompt-level reward statistics are reliable, but this assumption becomes fragile for binary, tied, skewed, or outlier-prone rewards. MAQ can be viewed as a robust alternative that maps each reward dimension to a common bounded normal-score scale, so the resulting advantages remain approximately comparable across dimensions while being less sensitive to pathological group statistics. Unlike a pure rank transformation, MAQ incorporates magnitude-aware gaps to preserve meaningful local quantitative differences among rollouts within the same prompt. Furthermore, unlike standard Z-score normalization, it compresses extreme gaps, thereby preventing a single outlier from dominating the prompt-level advantage allocation.

For each prompt $i$ and reward $k$, given a sorted group of $G$ rollout scores $r_{1} \le r_{2} \le \dots \le r_{G}$:

\begin{enumerate}
    \item \textbf{Log-compressed gaps:} We compute the spacing between adjacent rollouts:
    \begin{equation}
        gap_j = \log\left(1 + \frac{|r_{j+1} - r_{j}|}{\beta \cdot \sigma_{global}}\right)
    \end{equation}
    where $j = 1, \dots, G-1$. Here, $\sigma_{global}$ is the inter-quartile range (IQR) of reward $k$ across the global batch, serving as a robust scale baseline, and $\beta > 0$ controls the compression strength. This logarithmic compression is the key to robustness: it naturally restricts the influence of extreme outliers, while remaining approximately linear for small, dense gaps to preserve subtle intra-group distinctions.
    
    \item \textbf{CDF Allocation:} The gaps are normalized ($norm\_gap_j = gap_j / \sum_{j'=1}^{G-1} gap_{j'}$). The Cumulative Distribution Function (CDF) positions $u_{(j)}$ are then systematically allocated proportionally to these normalized gaps.
    
    \item \textbf{Inverse Normal Mapping:} Finally, the values are mapped to a standard normal distribution via the inverse CDF:
    \begin{equation}
        A_{(j)} = \Phi^{-1}(u_{(j)})
    \end{equation}
\end{enumerate}

As visualized in Figure~\ref{maq_compare}, MAQ reduces prompt-level advantage domination across the four task subspaces and maintains high effective rollout participation after normalization. Its role is therefore not to decorrelate reward dimensions directly, but to produce a more stable and comparable set of per-reward advantages before aggregation. This supports the projection-efficiency objective captured by $\eta_{\text{proj}}$ while leaving the remaining correlation redundancy to the whitening stage.

\subsubsection{Mahalanobis Whitening}

\begin{figure}[ht]
  \centering
  \includegraphics[width=0.95\textwidth]{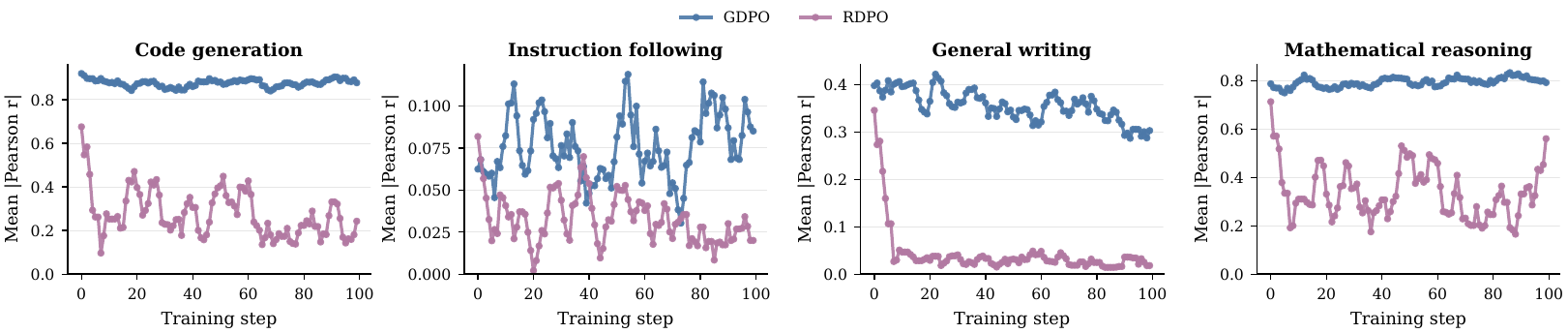}
  \caption{\textbf{Reward correlation within active task subspaces.} Each panel corresponds to one task-conditioned reward subspace. We report the mean absolute Pearson correlation among co-occurring reward dimensions over the course of training. Compared to GDPO, RDPO reduces within-subspace reward correlations by applying Mahalanobis whitening after MAQ normalization. This helps reduce the redundancy captured by $\eta_{\text{corr}}$ across the four-task training mixture.}
  \label{whiten_plot}
\end{figure}

\textbf{The Problem:} Although MAQ stabilizes individual reward dimensions at the prompt level, it does not inherently make different reward dimensions independent. This limitation is precisely what $\eta_{\text{corr}}$ measures: if two co-occurring rewards contain overlapping information, summing them can double-count the same variation; conversely, if they are negatively correlated, summing them can cancel out useful variation. In our current four-task mixture, such dependencies naturally arise within active reward subspaces. For instance, \texttt{math reward} or \texttt{code reward} can correlate with \texttt{length reward}, \texttt{ifeval reward} can correlate with \texttt{rubrics reward}, and \texttt{rm reward} can correlate with both \texttt{rubrics reward} and \texttt{length reward}; further details are provided in Section \ref{sec:training_setup}. Figure~\ref{whiten_plot} shows that these correlations are non-negligible under GDPO, particularly within the code generation, general writing, and mathematical reasoning subspaces.

\textbf{The Solution:} To mitigate the redundancy caused by inter-reward correlations, RDPO applies \textbf{Mahalanobis Whitening} following MAQ normalization. After MAQ, each rollout $(i, j)$ is represented by the advantage vector $\mathbf{A}^{(i,j)} = (A_1^{(i,j)}, A_2^{(i,j)}, \dots, A_n^{(i,j)})^T \in \mathbb{R}^n$. The whitening transformation maps this to a decorrelated vector:
\begin{equation}
\mathbf{W}^{(i,j)} = \hat{\boldsymbol{\Sigma}}_t^{-1/2}, \mathbf{A}^{(i,j)}
\label{eq:whitening}
\end{equation}
where $\hat{\boldsymbol{\Sigma}}_t^{-1/2} = \mathbf{U}\boldsymbol{\Lambda}^{-1/2}\mathbf{U}^T$ is computed via the eigendecomposition of the running covariance estimate $\hat{\boldsymbol{\Sigma}}_t = \mathbf{U}\boldsymbol{\Lambda}\mathbf{U}^T$. Given an accurate covariance estimate, this transformation targets $\mathrm{Cov}(\mathbf{W}) \approx \mathbf{I}_n$, shifting the active reward dimensions toward uncorrelated, unit-variance signals.

\textbf{Running Covariance Estimation.} During online RL training, the true reward covariance $\boldsymbol{\Sigma}$ is unknown and continuously shifts as the policy evolves. We maintain a stable estimate using an Exponential Moving Average (EMA) over the training steps:
\begin{equation}
\hat{\boldsymbol{\Sigma}}_t = (1 - \alpha)\,\hat{\boldsymbol{\Sigma}}_{t-1} + \alpha\,\hat{\boldsymbol{\Sigma}}_{\mathrm{batch}}
\label{eq:ema}
\end{equation}
where $\hat{\boldsymbol{\Sigma}}_{\mathrm{batch}}$ is the sample covariance computed from the current mini-batch of MAQ-normalized advantages, and $\alpha \in (0,1)$ is the EMA decay rate. The EMA smooths out batch-level noise and enables the whitening matrix to track the slowly evolving reward correlation structure. To ensure a reliable covariance estimate before applying the transformation, whitening begins only after a warmup phase of $T_{\mathrm{warm}}$ steps; in our implementation, we use the first five training steps for this warmup.

\textbf{Subspace Whitening for Heterogeneous Tasks.} In multi-task settings, a single rollout rarely observes all $n$ reward dimensions simultaneously. Our current training mixture consists of four active reward subspaces: \{\texttt{math}, \texttt{length}\}, \{\texttt{code}, \texttt{length}\}, \{\texttt{ifeval}, \texttt{rubrics}\}, and \{\texttt{length}, \texttt{rm}, \texttt{rubrics}\}. To accommodate this heterogeneity, we apply whitening exclusively over the \textit{observed subspace}: for a rollout with an active reward set $\mathcal{S} \subseteq \{1, \dots, n\}$, we extract the principal submatrix $\hat{\boldsymbol{\Sigma}}_{\mathcal{S}} \in \mathbb{R}^{|\mathcal{S}| \times |\mathcal{S}|}$ and compute $\hat{\boldsymbol{\Sigma}}_{\mathcal{S}}^{-1/2}$ independently. This approach ensures that decorrelation is applied only when reward dimensions co-occur within the same task, avoiding the introduction of artificial covariance estimates between dimensions that never overlap.

\textbf{Final Advantage.} The scalar advantage used for the PPO/GRPO policy gradient update is obtained by summing the whitened dimensions:
\begin{equation}
A_{\mathrm{sum}}^{(i,j)} = \sum_{k=1}^{n} W_k^{(i,j)} = \mathbf{1}^T \mathbf{W}^{(i,j)} = \mathbf{1}^T \hat{\boldsymbol{\Sigma}}_t^{-1/2}\,\mathbf{A}^{(i,j)}
\label{eq:final_adv}
\end{equation}

Under an ideal covariance estimate where $\mathrm{Cov}(\mathbf{W})=\mathbf{I}_n$, this projection captures less redundant information across dimensions. Because the covariance is estimated online via EMA and applied within specific observed task subspaces, this whitening process serves as a practical mechanism to reduce correlation redundancy rather than a strict mathematical guarantee of perfect decorrelation. The empirical curves in Figure~\ref{whiten_plot} show that this mechanism lowers the mean absolute reward correlation relative to GDPO in our training mixture. Combined with MAQ, this approach shifts the aggregated advantage toward a less redundant reward regime with higher effective information efficiency. As in GDPO, we subsequently apply batch-wise normalization to obtain the final advantage estimates.

\section{Training}

\subsection{Training Setup}
\label{sec:training_setup}

We apply RDPO during the post-training stage of LongCat-Flash. The policy is optimized on a four-task mixture comprising mathematical reasoning, code generation, instruction following, and general writing prompts. For each prompt, the model samples a set of rollouts, receives the specific subset of reward signals defined for that task, and constructs a scalar advantage from the active reward dimensions. Unless stated otherwise, the main model employs the complete RDPO pipeline. Specifically, MAQ normalization is first applied independently to each reward dimension to stabilize prompt-level advantage allocation. Mahalanobis whitening is then performed on the observed reward subspace to reduce correlation redundancy. Finally, the resulting whitened advantages are summed and batch-normalized prior to the policy-gradient update.

These four task categories activate distinct reward subspaces: mathematical reasoning samples use \texttt{math}+\texttt{length}, code generation samples use \texttt{code}+\texttt{length}, instruction-following samples use \texttt{ifeval}+\texttt{rubrics}, and general writing samples use \texttt{length}+\texttt{rm}+\texttt{rubrics}. Detailed descriptions of each reward are provided in the following section. This heterogeneous setting represents the intended use case for RDPO. Because varying tasks expose different reward subsets, the active rewards can differ substantially in scale, distributional shape, and correlation structure.

\begin{table}[htbp]
    \centering
    \caption{\textbf{Small-scale validation on a same-family smaller model.} We compare RDPO with GDPO, GRPO, and the RL initialization model on representative metrics.}
    \label{tab:internal-ckpt-selected-idpo-base-grpo-init}
    \tiny
    \setlength{\tabcolsep}{3pt}
    \resizebox{0.92\textwidth}{!}{
    \begin{tabular}{@{}lccccccc@{}}
      \toprule
      \textbf{Method} & \textbf{IFEval} & \textbf{AIME24} & \textbf{AH-Hard} & \textbf{AH-Creative} & \textbf{FullStack} & \textbf{HumanEval+} & \textbf{MBPP+} \\
      \midrule
      Init. & 72.64 & 59.58 & 12.90 & \textbf{30.10} & 55.57 & 82.93 & 77.84 \\
      GRPO & 76.52 & 60.21 & 13.30 & 26.10 & 55.87 & 82.93 & 77.78 \\
      GDPO & 78.56 & 58.54 & 14.00 & 24.00 & 55.87 & 83.54 & 77.25 \\
      RDPO & \textbf{83.55} & \textbf{60.42} & \textbf{14.40} & 21.50 & \textbf{56.91} & \textbf{85.37} & \textbf{78.51} \\
      \bottomrule
    \end{tabular}}
\end{table}

\begin{table}[htbp]
    \centering
    \caption{\textbf{Component validation on the same smaller-model setting.} We use the same representative benchmark set to compare the base GDPO setup, MAQ-only (\textbf{Q}), whitening-only (\textbf{M}), and the combined RDPO variant (\textbf{Q+M}).}
    \label{tab:internal-ckpt-selected-qm-q-m-base}
    \tiny
    \setlength{\tabcolsep}{3pt}
    \resizebox{0.92\textwidth}{!}{
    \begin{tabular}{@{}lccccccc@{}}
      \toprule
      \textbf{Method} & \textbf{IFEval} & \textbf{AIME24} & \textbf{AH-Hard} & \textbf{AH-Creative} & \textbf{FullStack} & \textbf{HumanEval+} & \textbf{MBPP+} \\
      \midrule
      Base & 78.56 & 58.54 & 14.00 & 24.00 & 55.87 & 83.54 & 77.25 \\
      Q    & 81.70 & 59.69 & 12.30 & \textbf{27.50} & 56.19 & \textbf{85.37} & 77.78 \\
      M    & 81.33 & 60.31 & 13.80 & 25.20 & 56.70 & 84.15 & 77.25 \\
      Q+M  & \textbf{83.55} & \textbf{60.42} & \textbf{14.40} & 21.50 & \textbf{56.91} & \textbf{85.37} & \textbf{78.51} \\
      \bottomrule
    \end{tabular}}
  \end{table}

\subsection{Reward Design}
\label{sec:reward_design}

\textbf{Rubrics Reward} For each sampled response, we perform fine-grained validation against its associated rubric set using a generative reward model. The evaluation result for each rubric is recorded as a binary variable. We then compute a weighted average using predefined rubric weights to obtain the final rubric reward. If a response fails any criterion marked as essential, the total rubric reward is strictly set to $0$. Otherwise, we compute a normalized weighted sum over all valid rubrics and clip the result to the $[0, 1]$ interval. This design ensures the reward captures both broad coverage of explicit writing requirements and strict satisfaction of critical constraints.

\textbf{IFEval Reward} The IFEval reward measures whether a response adheres to explicit instruction constraints. For each response, we invoke a rule-based verifier associated with the reference annotation to evaluate formatting, content, or behavioral requirements. While standard IFEval annotations yield a strict pass/fail signal, certain extended datasets provide continuous scores. In both scenarios, this reward offers direct supervision for instruction-following capabilities and primarily reflects compliance with hard task constraints.

\textbf{Math Reward} The math reward evaluates the correctness of mathematical reasoning. For problems with verifiable final answers, the grader extracts the generated answer and compares it against the reference solution using exact-match or task-specific equivalence checks. This metric provides the primary correctness signal for mathematical samples, while the length reward applies complementary pressure toward concise reasoning.

\textbf{Code Reward} The code reward assesses the functional correctness of generated programs. For coding tasks, the grader evaluates the generated solution using a reference evaluation protocol, such as execution-based checks or task-specific validators when available. This reward is paired with the length reward to ensure that code-oriented reinforcement learning optimizes both correctness and response efficiency.

\textbf{RM Reward} The RM reward is generated by an independent reward model to capture holistic response quality. We concatenate the prompt and response into a complete dialogue and feed it into the reward model to obtain a raw scalar score. Because these raw outputs can span a wide range, we linearly rescale the scores to $[0, 1]$ to maintain numerical consistency with other reward components. Unlike rule-based metrics such as rubrics and IFEval, the RM reward provides a soft preference signal for fluency, completeness, coherence, and subjective quality. Consequently, it serves as a complementary signal rather than a substitute for hard task constraints.

\textbf{Length Reward} The length reward encourages concise responses without compromising task satisfaction. For each response, the generated length is compared against a reference statistic: the average length of successful task completions across multiple samplings from the base model for a given query. This metric reflects the base model's inherent capability and establishes a robust baseline for subsequent training. Responses with a length below this threshold receive a reward of $1$. Conversely, if the length exceeds the threshold, the reward decays according to a quadratic penalty and is clipped to the $[0, 1]$ interval. This formulation avoids over-penalizing minor length overruns while imposing a stricter penalty on distinctly verbose generations.

\textbf{Conditional Reward Handling} Before combining multiple rewards, we apply a conditional handling mechanism to prevent auxiliary signals from compensating for failures in core requirements. The RM reward is constrained by the rubric reward; specifically, if the rubric reward falls below $0.5$, the RM reward is truncated to $\min(r_{\text{rubric}}, r_{\text{rm}})$. This ensures that a high holistic preference score cannot mask violations of essential rubrics. We apply an analogous gating rule to the length reward. For instruction-following samples, the length reward is considered valid only when the IFEval constraint is satisfied. If the IFEval score drops below $0.5$, the length reward is reduced accordingly. For math, code, and rubric-based writing samples, the length reward is similarly truncated when the primary task reward falls below $0.5$. Consequently, length control and holistic preference function as auxiliary optimization signals only when the response already satisfies the fundamental task requirements.

\section{Evaluation}

\begin{table}[htbp]
  \centering
  \caption{\textbf{Scaled LongCat-Flash post-training comparison.} We compare the RL initialization model (\textbf{Init.}), GRPO, and RDPO on representative benchmarks spanning instruction following, math and knowledge reasoning, writing and arena-style evaluation, and coding.}
  \label{tab:final-four-models-no-code}
  \footnotesize
  \setlength{\tabcolsep}{4pt}
  \begin{tabular}{@{}>{\raggedright\arraybackslash}p{11.8cm}ccc@{}}
    \toprule
    \textbf{Metric} & \textbf{Init.} & \textbf{GRPO} & \textbf{RDPO} \\
    \midrule
    \multicolumn{4}{@{}l}{\textit{Instruction Following}} \\
    \hline
    IFEval\textsubscript{Acc} & 86.14\% & 89.46\% & \textbf{90.39\%} \\
    GuideBench\textsubscript{Acc} & \textbf{87.81\%} & 84.35\% & 87.04\% \\
    SOP-Maze\textsubscript{Acc} & 37.80\% & 34.83\% & \textbf{38.17\%} \\
    \midrule
    \multicolumn{4}{@{}l}{\textit{Math and Knowledge Reasoning}} \\
    \hline
    AIME2024\textsubscript{Avg@32} & \textbf{86.3\%} & 85.56\% & 85.73\% \\
    AIME2025\textsubscript{Avg@32} & 78.31\% & 77.29\% & \textbf{78.85\%} \\
    GPQA\textsubscript{Avg@16} & 66.36\% & \textbf{68.54\%} & 67.79\% \\
    MATH500\textsubscript{Acc} & \textbf{98.60\%} & 98.20\% & \textbf{98.60\%} \\
    \midrule
    \multicolumn{4}{@{}l}{\textit{Writing and Arena Evaluation}} \\
    \hline
    WritingBench\textsubscript{Acc} & 83.17\% & 84.12\% & \textbf{87.63\%} \\
    ArenaHard-v2 (Creative)\textsubscript{Acc} & 70.60\% & 82.50\% & \textbf{89.00\%} \\
    ArenaHard-v2 (Hard)\textsubscript{Acc} & 49.40\% & 65.80\% & \textbf{76.10\%} \\
    \midrule
    \multicolumn{4}{@{}l}{\textit{Coding}} \\
    \hline
    FullStackBench\textsubscript{Pass@1} & 65.06\% & \textbf{67.16\%} & 66.48\% \\
    HumanEval+\textsubscript{Pass@1} & \textbf{92.68\%} & 89.02\% & 91.46\% \\
    MBPP+\textsubscript{Pass@1} & 78.84\% & 78.57\% & \textbf{79.63\%} \\
    LiveCodeBench(24.08-25.01)\textsubscript{Pass@1} & 63.08\% & 60.93\% & \textbf{63.80\%} \\
    \bottomrule
  \end{tabular}
\end{table}

\subsection{Evaluation Setup}

To evaluate performance across the trained task categories, we select a diverse set of challenging benchmarks and organize them into four evaluation clusters:

\begin{enumerate}
    \item \textbf{Instruction Following:} This cluster includes \textbf{IFEval}~\citep{ifeval}, \textbf{GuideBench}~\citep{guidebench}, and \textbf{SOP-Maze}~\citep{sop_maze}.

    \item \textbf{Math and Knowledge Reasoning:} This cluster includes \textbf{AIME24}, \textbf{AIME25}, \textbf{GPQA}~\citep{gpqa}, and \textbf{MATH500}~\citep{math500}.
    
    \item \textbf{Writing and Arena Evaluation:} This cluster includes \textbf{WritingBench}~\citep{WritingBench} and \textbf{ArenaHard v2}~\citep{arenahard_v2}. For ArenaHard v2, we report two complementary subsets: \textbf{AH-Hard} and \textbf{AH-Creative}.
    
    \item \textbf{Coding:} This cluster includes \textbf{FullStackBench}~\citep{fullstackbench}, \textbf{HumanEval+}~\citep{humaneval}, \textbf{MBPP+}~\citep{mbpp}, and \textbf{LiveCodeBench v6}~\citep{livecodebench}.
\end{enumerate}

\subsection{Small-Scale Validation on a Same-Family Smaller Model}

Prior to scaling RDPO to the larger LongCat-Flash post-training run, we first validate the method on a smaller internal model from the same family. This preliminary stage serves two primary purposes: evaluating whether the complete reward-decorrelated pipeline improves upon relevant baselines, and isolating the contributions of its two core components: MAQ normalization and Mahalanobis whitening.

Tables~\ref{tab:internal-ckpt-selected-idpo-base-grpo-init} and~\ref{tab:internal-ckpt-selected-qm-q-m-base} show encouraging preliminary performance, supporting a larger-scale LongCat-Flash trial. The full pipeline improves over the GDPO baseline across IFEval, AIME24, AH-Hard, FullStackBench, HumanEval+, and MBPP+. Furthermore, component-level analysis suggests that MAQ and whitening provide complementary benefits: MAQ is strong on several distribution-sensitive metrics, including AH-Creative, while whitening helps in correlation-sensitive settings. Together, these empirical results motivate adopting the complete RDPO recipe for the LongCat-Flash post-training run.

\subsection{Scaled LongCat-Flash Post-Training Results}

Following the small-scale validation phase, we scale the full RDPO pipeline to LongCat-Flash. Our LongCat-Flash evaluation focuses on end-to-end scalability. Specifically, we examine how the complete reward-decorrelated advantage construction behaves in a larger post-training regime.

As shown in Table~\ref{tab:final-four-models-no-code}, the LongCat-Flash RDPO model primarily yields gains in capabilities aligned with its mixed-reward training objective. Among the three models evaluated, RDPO attains the highest scores on IFEval and SOP-Maze, alongside distinct improvements on WritingBench and both reported ArenaHard v2 subsets (AH-Creative and AH-Hard). These results are consistent with our small-scale validation: stabilizing prompt-level advantage allocation and reducing reward redundancy appear useful for instruction-following and open-ended, preference-sensitive evaluations.

On the remaining reasoning and coding evaluations, the comparison is mixed but stable. RDPO matches the best MATH500 score and remains competitive on AIME2025 and GPQA, while Init. or GRPO can remain stronger on individual metrics. Coding results follow a similar pattern: RDPO leads on MBPP+ and the LiveCodeBench v6, while GRPO or Init. remains stronger on FullStackBench and HumanEval+. Overall, the scaled LongCat-Flash experiment suggests that the full RDPO recipe transfers from smaller-model validation with broadly stable reasoning and coding results.

\section{Conclusion}

RDPO combines prompt-level MAQ normalization with subspace-level Mahalanobis whitening to stabilize mixed-reward RL, improving LongCat-Flash post-training on instruction following, writing, and ArenaHard v2, with broadly competitive results on reasoning and coding evaluations.

\section{Acknowledgement}

We sincerely thank the infrastructure team and evaluation team of LongCat for their constructive feedback and prompt support.

\bibliographystyle{unsrtnat}
\bibliography{main}

\end{document}